\documentclass{article} 
\usepackage{iclr2025_conference,times}

\usepackage{amsmath,amsfonts,bm}

\def\eqref#1{equation~\ref{#1}}

\def\1{\bm{1}}

\DeclareMathAlphabet{\mathsfit}{\encodingdefault}{\sfdefault}{m}{sl}
\SetMathAlphabet{\mathsfit}{bold}{\encodingdefault}{\sfdefault}{bx}{n}

\usepackage{hyperref}
\usepackage{url}

\usepackage{amsmath}
\usepackage{amsthm}
\usepackage{amssymb}
\usepackage{microtype}

\usepackage{multirow}
\usepackage{booktabs}
\usepackage{graphicx}
\title{To Retrieve or Not to Retrieve? Uncertainty Detection for Dynamic Retrieval Augmented Generation}

\author{Kaustubh D. Dhole  \\
Department of Computer Science\\
Emory University\\
Atlanta, GA 30307, USA \\
\texttt{kdhole@emory.edu} \\
}

\iclrfinalcopy 
\begin{document}

\maketitle

\begin{abstract}
Retrieval-Augmented Generation equips large language models with the capability to retrieve external knowledge, thereby mitigating hallucinations by incorporating information beyond the model's intrinsic abilities. However, most prior works have focused on invoking retrieval deterministically, which makes it unsuitable for tasks such as long-form question answering. Instead, dynamically performing retrieval by invoking it only when the underlying LLM lacks the required knowledge can be more efficient. In this context, we delve deeper into the question, ``To Retrieve or Not to Retrieve?'' by exploring multiple uncertainty detection methods. We evaluate these methods for the task of long-form question answering, employing dynamic retrieval, and present our comparisons. Our findings suggest that uncertainty detection metrics, such as Degree Matrix Jaccard and Eccentricity, can reduce the number of retrieval calls by almost half, with only a slight reduction in question-answering accuracy.

\end{abstract}

\section{Introduction}
Recently, Large Language Models (LLMs) like ChatGPT~\cite{openai2023gpt4}, Gemini~\cite{team2023gemini}, and others are showing impressive strides in tasks across numerous benchmarks~\cite{srivastava2023beyond}. This success has been largely owed to their exposure to massive training data and successive fine-tuning of instruction datasets. To increase the helpfulness and decrease the harmfulness of the models, they are being further fine-tuned over preference collections~\cite{bai2022training, rlhf, rafailov2024direct}. 

Further, Retrieval Augmented Generation (RAG)~\cite{lewis2020retrieval,dhole2024kaucus,dhole2024conqret}, in the effort to mitigate hallucinations, enriches these models with domain-specific information and tackles scenarios where the intrinsic knowledge of the base model falls short. By integrating externally retrieved content during the generation phase, RAG enhances the model's ability to produce less hallucinatory and domain-conditioned responses. This approach has been particularly valuable in complex applications such as long-form generation like multi-hop question answering, which often requires multiple retrievals to address a query comprehensively.

However, to optimize the efficiency of RAG, retrieval should only be invoked when necessary --- also referred to as conditional retrieval. Previous conditional RAG setups have explored multiple paradigms like low token probabilities~\cite{jiang-etal-2023-active}, external classifiers~\cite{wang-etal-2023-self-knowledge}, or low entity popularity~\cite{mallen-etal-2023-trust} as indicators of the LLMs' knowledge gaps. However, most of these methods fall short in either approximating knowledge gaps of the LLMs or lacking the ability to invoke retrieval dynamically.

On the other hand, with the potential of LLMs to hallucinate, there has been an increasing interest in~\textbf{uncertainty detection} methods to gauge LLMs' confidence in their outputs~\cite{fadeeva-etal-2023-lm}. Unlike traditional methods that rely on rigid heuristics or external classifiers, uncertainty detection leverages the inherent variability in LLM-generated responses to estimate confidence dynamically.

For instance, semantic sets-based UD approaches~\cite{lingenerating} group responses based on meaning, and use the number of clusters to directly reflect the level of uncertainty --- with greater variability signaling higher uncertainty. Similarly, spectral methods using eigenvalue Laplacians quantify response diversity by identifying strong or weak clustering patterns in pairwise similarity graphs. These approaches align with the probabilistic nature of LLMs as well as adaptively gauge uncertainty based on output coherence, making them more robust against adversarial or ambiguous inputs. 

In this work, we evaluate if such uncertainty detection methods can indeed enhance the reliability of conditionally invoking retrieval, by measuring its impact on a downstream task of multi-hop question answering.

In that regard, we resort to a conditional RAG system and employ numerous uncertainty detection metrics to test the need for invoking retrieval. Our RAG system performs forward-looking active retrieval in the style of~\citet{jiang-etal-2023-active}.

Specifically, we contribute the following:
\begin{itemize}
    \item We design a method that performs retrieval augmented generation with dynamic retrieval through uncertainty detection
    \item We perform an exhaustive analysis of various conditions from the ``uncertainty quantification'' literature to gauge the best strategy to dynamically retrieve during generation
    \item Based on the results, we present insights for future research
\end{itemize}

Our insights are useful to gauge whether uncertainty detection methods can help improve the efficiency of RAG.

\section{Related Work}

Here, we summarise some of the related work on uncertainty quantification and some active RAG efforts.

There has been a lot of recent work on uncertainty quantification of white box and black box NLG models.~\citet{lingenerating} showed that along with their generations, GPT-3 can output a verbalized form of the uncertainty, viz. ``high confidence'' or ``85\% confidence''.~\citet{kadavath2022language} show that models can be made to sample answers and then made to self-evaluate the probability of P(True).~\citet{kuhn2023semantic} recently proposed to compute the semantic entropy by considering the equivalence relationships amongst generated responses.

\cite{wang2024self} proposed Self-DC that tackled compositional questions via iterative divide-and-conquer based on LLM certainty.~\cite{yao2024seakr} propose utilising the model's internal states to estimate uncertainty and deciding whether to retrieve or not.

We now describe the tasks and datasets used in our analysis along with the UD approaches employed.
\section{Tasks and Datasets}
We conduct experiments on the 2WikiMultihopQA dataset~\cite{xanh2020_2wikimultihop}, a multi-hop open domain question answering (QA) dataset that tests the reasoning and inference skills of question-answering models. Questions in this dataset generally require two steps of reasoning to deduce the final answer, and the information for each step of reasoning can be obtained through referencing external information viz., Wikipedia passages.

\section{Approach}
We now describe our uncertainty-aware, retrieval-augmented generation in the following two subsections.
\subsection{Uncertainty Evaluation of Future Sentence}
Given a query $\mathbf{q}$, a retriever $\mathbf{R}$, a text generator $\mathbf{G}$, and a black box uncertainty estimation function $\mathbf{U}$, and partially generated sequence $t_{<i}$ until time step $i$, -- we first generate a temporary sentence $t_n$ in the style of FLARE~\cite{jiang-etal-2023-active}. 

We use a prompt template $\mathbf{P}$, which could take the form of a zero-shot or a few-shot instruction. This instruction takes as input the query, zero or more retrieved documents $d_1 \ldots d_k$, and the answer tokens generated until now. Here, we use $t_{i}$ to represent the $i^{th}$ temporary sentence and $y_{<i}$ to represent all the initialised and generated sentences $\{0 \ldots (i-1)\}$. $t_{i}$ is first obtained without performing retrieval:
\begin{equation}
    t_{i} = \mathbf{G}(\mathbf{P}\{\mathbf{q}, \ldots, y_{i-1}\})
\end{equation}

During generation, we evaluate the uncertainty of this temporary sentence $t_n$ to gauge if the generator needs more information. If the uncertainty $\mathbf{U}(t_n)$ exceeds a threshold $\theta_{\mathbf{U}}$, the model is not certain and may lack the necessary knowledge to provide an accurate answer. The next sentence $y_{i}$ is then computed by appending retrieved information to the model context:
\begin{equation}
y_{i} = 
\begin{cases} 
\mathbf{G}(\mathbf{P}\{d_1, \ldots, d_k, \mathbf{q}, \ldots, y_{i-1}\}) & \text{if}~\mathbf{U}(t_i) > \theta_{\mathbf{U}}\\
\mathbf{G}(\mathbf{P}\{\mathbf{q}, \ldots, y_{i-1}\}) & otherwise
\end{cases}
\end{equation}
where $d_1 \ldots d_k$ are obtained from a retrieval system $\mathbf{R}$.
\begin{equation}
d_1 \ldots d_k := \mathbf{R} (\mathbf{q})
\end{equation}

\subsection{Sequence Level Uncertainty Evaluation Measures}
We resort to 5 recently introduced sequence-level uncertainty evaluation measures. Each of them work in a black box manner without requiring information regarding the model parameters. 

The high-level strategy of all the methods is the same. Given an input $x$, first generate $n$ responses through some generator $G$ and then compute pairwise similarity scores of each of the $n$ responses with each other. Using these similarity values, compute an uncertainty estimate $U(x)$ or a confidence score.
\begin{itemize}
    \item \textbf{Semantic Sets}: In the black-box approach of~\cite{kuhn2023semantic}, the authors propose to compute semantic sets i.e. groups of responses that are close together in meaning. These semantic sets of equivalence subsets are computed using a Natural Language Inference (NLI) classifier. Here, the number of semantic sets can be regarded as an uncertainty estimate as when the responses differ in meaning, the number of groups increases. 
    \item \textbf{Eigen Value Laplacian}: defines the uncertainty estimate by capturing the essence of spectral clustering. First, an adjacency matrix is created from the pairwise similarities of responses. Then the matrix is partitioned into clusters, where each cluster corresponds to a distinct ``meaning'' or category within the responses. The eigenvalues close to one indicate strong cluster formations, thus contributing less to the uncertainty estimate, while those further from one suggest weaker clustering or more diffuse distributions of responses, hence increasing the uncertainty estimate.
    \\
    The degree matrix of the adjacency graph is also used to compute the uncertainty estimate~\cite{lingenerating}. A node that is well-connected to other nodes, might be less uncertain. We use two similarity metrics for computing the degree matrix.
    \item \textbf{Degree Matrix (Jaccard Index)}:  The Jaccard similarity is a light-weight metric where sentences or passages are treated as sets of words, and similarity between responses is computed by taking the fraction of the intersection of the two sets and the union of the two sets.
    \item \textbf{Degree Matrix (NLI)}: Here, the similarity between responses is computed through classifying entailment relations amongst them. A classifier predicts whether a pair of responses contradict, entail, or are neutral to each other.
\end{itemize}
\begin{table*}[!htpb]
\centering
\resizebox{\textwidth}{!}{%
\begin{tabular}{lc|ccccc}
\toprule
\textbf{Uncertainty Estimator} & \textbf{Trigger Retrieval When} & \textbf{Retrieval Query} & \textbf{\#examples} & \textbf{\#search} & \textbf{\#steps} & \textbf{f1} \\
\cmidrule{1-2} \cmidrule{3-7}
Always Retrieve & U $\geq$ 0 & Temporary Sentence & 25 & 4.60 & 3.60 & 0.552 \\
Always Retrieve&  & Sub-Query & 25 & 5.00 & 4.00 & 0.538 \\
\hline
FLARE-Instruct & ``...[Search'' &  & 25 & 4.80 & 3.80 & 0.531 \\
Degree Matrix Jaccard & U $>$ 0.4 & Sub-Query & 24 & 1.46 & 3.67 & 0.593 \\
Eccentricity & U $>$ 2 & Sub-Query & 22 & 2.23 & 4.05 &~\textbf{0.605} \\
Semantic Sets & U $>$ 2 & Sub-Query & 23 & 2.52 & 4.09 & 0.411 \\
Degree Matrix NLI & U $>$ 0.5 & Sub-Query & 24 & 2.25 & 4.00 & 0.535 \\
\bottomrule
\end{tabular}}
\caption{Performance Metrics over a smaller seed set}
\label{perf1}
\end{table*}

\begin{table*}[!htpb]
\centering
\resizebox{\textwidth}{!}{%
\begin{tabular}{lc|ccccccc}
\toprule
\textbf{Uncertainty Estimator} & \textbf{Trigger Retrieval When} & \textbf{\#search} & \textbf{\#steps} & \textbf{ret ratio} & \textbf{correct} & \textbf{incorrect} & \textbf{f1} \\
\cmidrule{1-2} \cmidrule{3-8}
Always Retrieve & Always & 4.63 & 3.63 & 1.32 & 0.493 & 0.493 & 0.578 \\
 &  & 4.61 & 3.61 & 1.33 & 0.52 & 0.467 & 0.594 \\
 &  & 4.61 & 3.61 & 1.33 & 0.493 & 0.493 & 0.571 \\
 \cmidrule{3-8}
 &  &  &  &  &  &  & \textbf{0.581} \\
\cmidrule{1-8}
Degree Matrix Jaccard & U $>$ 0.4 & 1.80 & 3.61 & 0.57 & 0.453 & 0.533 & 0.538 \\
 &  & 1.92 & 3.60 & 0.61 & 0.44 & 0.547 & 0.525 \\
 &  & 1.85 & 3.61 & 0.57 & 0.419 & 0.568 & 0.508 \\
 \cmidrule{3-8}
 &  &  &  &  &  &  & \textbf{0.524} \\
\cmidrule{1-8}
Eccentricity & U $>$ 2 & 2.17 & 3.60 & 0.64 & 0.44 & 0.547 & 0.525 \\
 &  & 2.25 & 3.63 & 0.67 & 0.467 & 0.533 & 0.565 \\
 &  & 2.23 & 3.63 & 0.64 & 0.507 & 0.493 & 0.594 \\
 \cmidrule{3-8}
 &  &  &  &  &  &  & \textbf{0.561} \\
\bottomrule
\end{tabular}}
\caption{Performance Metrics for Different Uncertainty Estimators for 75 examples.}
\label{perf2}
\end{table*}
\subsection{Subquery Generation for Retrieval}
We resort to retrieving relevant knowledge to account for the information that the model is lacking to answer the question. FLARE~\cite{jiang-etal-2023-active} generates a retrieval query for the missing entity in the temporary sentence by using the sentence with the low probability token removed or by prompting an external question generator to generate a question for the missing entity as the answer. We generalize this by instead prompting the model to generate a subquery to figure out the missing information needed to answer the user query in an open-ended manner.

We define a subquery generator $\mathbf{S_Q}$ which takes in as input few-shot exemplars of subqueries, the current user query $\mathbf{q}$, and the current partial answer sentences uttered in chain-of-thought~\cite{wei2022chain} fashion. It seeks to generate subqueries to get a specific piece of information not generated in the partial answer sentences but is needed to answer $\mathbf{q}$. Once this subquery is generated, we use this subquery to retrieve additional passages from the external retriever $\mathbf{R}$. These passages are then appended to the user input, and the generation continues. 

For instance, for the question, ``Which film has the director who died first, Promised Heaven or Fire Over England?'', and the partially generated answer, ``The film Promised Heaven was directed by Eldar Ryazanov. Fire Over England was directed by William K. Howard. Eldar Ryazanov died on November 30, 2015.'', we expect the model to generate a subquery, ``When did William K. Howard die?''.

\section{Setup} The generator used in all experiments was GPT-3 (davinci-002)~\cite{brown2020language}, and the retriever employed was BM25 through PyTerrier~\cite{macdonald2021pyterrier, dhole2024pyterrier}. The base code used for conducting the experiments and computing the metrics presented in the tables was obtained from the active RAG setup by~\citet{jiang-etal-2023-active}. For uncertainty detection, we resort to the~\citet{fadeeva-etal-2023-lm}'s LM-Polygraph library.

Since running GPT-3 (davinci-002) along with many of the uncertainty detection metrics could be expensive to run (due to making multiple calls), we first perform a run for a small seed set of 25 queries across all metrics and then choose the 3 best metrics for a rerun across a larger set of 75 examples. We perform each run three times.

\section{Results}

We now present the results in Tables~\ref{perf1} and~\ref{perf2} for the smaller and the larger sets respectively. 

The baseline method where retrieval was always invoked yielded an F1 score of~\textbf{0.552} when using temporary sentences as retrieval queries and~\textbf{0.538} when subqueries were generated for retrieval but required most number of retrieval operations.

Triggering retrieval, when uncertainty computed through~\textbf{Eccentricity} i.e. \textbf{$U > 2$}, led to the highest F1 score of~\textbf{0.605}, with a lesser number of search operations. This approach balanced retrieval efficiency and task performance better than other methods. It required half the number of search operations than an Always Retrieve approach.~\textbf{Semantic Sets'} innovative clustering approach performed poorly, with an F1 score of~\textbf{0.411}. Using entailment-based similarity to compute uncertainty via the \textbf{Degree Matrix NLI} measure achieved an F1 score of~\textbf{0.535}, comparable to the baseline. The lightweight~\textbf{Degree Matrix (Jaccard)} necessitated the least number of retrieval operations to perform better than an Always Retrieve baseline.

Table~\ref{perf2} presents additional performance metrics over a larger set of 75 examples. Notably, the \textbf{Eccentricity} method consistently demonstrated the best balance between retrieval efficiency and performance, achieving an average F1 score of \textbf{0.561} across different experimental runs, while reducing unnecessary retrievals compared to the baseline.

\textbf{Degree Matrix (Jaccard)} performed slightly worse in F1 score (\textbf{0.524}) but depended on retrieval the least indicating its potential for applications where minimizing retrieval costs is crucial.

In contrast, the~\textbf{Always Retrieve} approach performed better than both conditional retrieval approaches but necessitated almost twice the number of retrieval calls.

\section{Conclusion}
Our experiments demonstrate that dynamic retrieval, guided by uncertainty detection, improves the efficiency of retrieval-augmented generation systems, making it useful where retrieval can be expensive to compute. Among the methods tested,~\textbf{Eccentricity-based uncertainty detection} emerged as the best-performing approach, offering the highest F1 score with a moderate number of retrieval steps and searches. This method effectively balances retrieval efficiency with task performance.

The~\textbf{Degree Matrix (Jaccard)} method also showed promising results, particularly in reducing retrieval costs while maintaining reasonable performance. Conversely, methods such as~\textbf{Semantic Sets} and~\textbf{FLARE-Instruct} underperformed, highlighting the need for more reliable uncertainty estimators.

Although some black-box uncertainty detection methods require multiple runs of generation, which can be costly, always retrieving may be preferable in RAG applications where lightweight retrieval methods like BM25 suffice. This is also evident from the results on the larger set. 


Besides, we feel that uncertainty detection might become more mainstream as the propensity for hallucination in LLMs increases, and as end applications demand more confidence and interpretability~\cite{dhole2024conqret} in their outputs making uncertainty detection a necessity. Our work focuses on exploiting uncertainty detection for RAG, especially where retrieval can be expensive like the usage of heavy and composite retrieval systems employing numerous components like reformulation, dense retrieval~\cite{santhanam2021colbertv2}, reranking, etc.
\section{Ethical Considerations}

When evaluating large language models (LLMs), it is essential to adopt a sociotechnical perspective~\cite{dhole2023large}, acknowledging that their outputs are influenced by both social contexts and technical design choices. Proper safeguards should be in place to mitigate biases and prevent the generation of harmful or toxic content. Furthermore, the uncertainty detection approaches we employed rely on estimations derived from various neural network computations, which are inherently shaped by the data on which the models are trained. Consequently, it is critical to thoroughly test uncertainty detection methods to ensure they meet the requirements of the intended applications.

Despite these precautions, there remains a possibility that some approaches may misrepresent the level of certainty, as no method is flawless. Therefore, ongoing evaluation and refinement of uncertainty detection mechanisms are necessary to minimize inaccuracies and potential misinterpretations.

\section*{Acknowledgements}
The author would like to thank Eugene Agichtein for insightful discussions and the anonymous reviewers for their useful feedback. The author would also like to thank Microsoft for providing OpenAI credits through the Microsoft Accelerating Foundation Models Research Award.

\bibliography{iclr2025_conference}
\bibliographystyle{iclr2025_conference}

\end{document}